\documentclass{article}
\usepackage{spconf,amsmath,graphicx,hyperref}
\usepackage{microtype}
\usepackage{graphicx}
\usepackage{subfigure}
\usepackage{booktabs} 
\usepackage{multirow}
\usepackage{array}
\usepackage{hyperref}
\usepackage{makecell}
\newcolumntype{H}{>{\setbox0=\hbox\bgroup}c<{\egroup}@{}}
\usepackage{tabularx}

\title{Robust MAE-Driven NAS: From Mask Reconstruction to Architecture Innovation}
\name{Yiming Hu, Xiangxiang Chu, Yong Wang}
\address{AMAP, Alibaba Group}

\begin{document}
%
\maketitle
\begin{abstract}
Neural Architecture Search (NAS) relies heavily on labeled data, which is labor-intensive and time-consuming to obtain. In this paper, we propose a novel NAS method based on an unsupervised paradigm, specifically Masked Autoencoders (MAE), thereby eliminating the need for labeled data. By replacing the supervised learning objective with an image reconstruction task, our approach enables the efficient discovery of network architectures without compromising performance and generalization ability. Additionally, we address the problem of performance collapse encountered in the widely-used Differentiable Architecture Search (DARTS) in the unsupervised setting by designing a hierarchical decoder. Extensive experiments across various datasets demonstrate the effectiveness and robustness of our method, offering empirical evidence of its superiority over the counterparts.
\end{abstract}

\section{Introduction}
\label{sec:intro}

Neural Architecture Search (NAS) has attracted significant attention in recent years, emerging as a crucial advancement in the machine learning domain. This surge of interest~\cite{zela2020understanding,liang2019darts,ramachandran2017searching,liu2019auto} is largely due to NAS algorithms' ability to autonomously discover superior architectures, thereby expediting the design process and alleviating the reliance on human expertise. These algorithms have demonstrated excellent performance in areas such as image classification, object detection, and more.

Traditionally, existing NAS methods have concentrated on leveraging supervised learning frameworks, requiring extensive labeled datasets to guide the search for optimal architectures. While effective, this approach is inherently limited by the considerable resources needed to generate large-scale annotated datasets. Recognizing this challenge, recent research~\cite{liu2020labels,yan2020does,zhang2021neural} has begun exploring avenues to reduce dependence on such data, focusing on more efficient search strategies that do not compromise performance.

\begin{figure*}[t]
  \centering
  \includegraphics[width=0.88\textwidth]{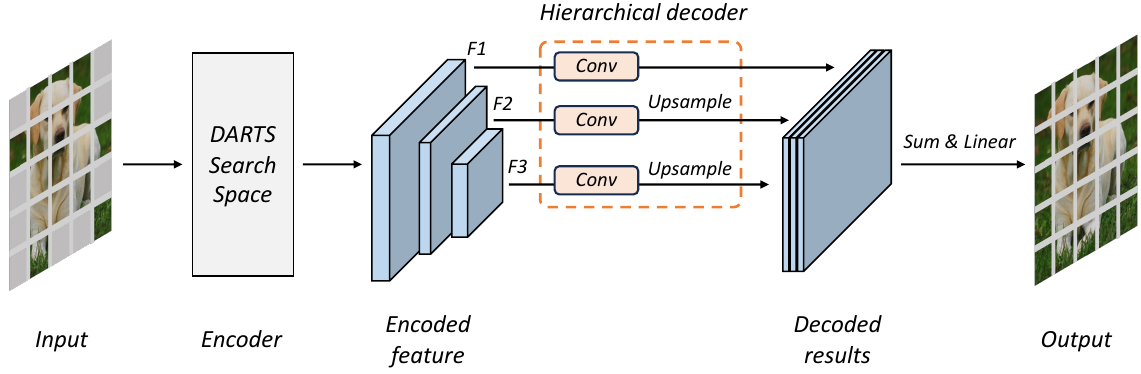}
  \caption{The framework of MAE-NAS. The input is an image with an applied mask, which is first fed into an encoder and then passed a hierarchical decoder, ultimately producing a reconstructed image. The encoder section is built on the search space for NAS, aimed at selecting a superior architecture to enhance the quality of the reconstructed image.}
  \label{fig:mae-nas}
\end{figure*}

In this study, we introduce a novel NAS method called MAE-NAS, which leverages the capabilities of Masked Autoencoders~\cite{he2022masked}. To our knowledge, this represents an innovative exploration, as previous research has only minimally explored this area. Our approach applies the unsupervised methodology of masked image modeling~\cite{xie2022simmim, he2022masked} to the well-established DARTS framework~\cite{liu2018darts}. By replacing the supervised objective of NAS with an image reconstruction task, our method has more efficient search phase and eliminates the need for labeled data. As depicted in Figure~\ref{fig:mae-nas}, our framework processes randomly masked images through an encoder, which is built on the DARTS search space, and utilizes a decoder to reconstruct the original image. In particular, we address the performance collapse issue of DARTS in unsupervised settings, identifying a strong correlation between collapse occurrence and the masking ratio. Specifically, a higher mask ratio (e.g., greater than 0.5) enables DARTS to mitigate performance collapse effectively. Building on this insight, we propose a hierarchical decoder to stabilize the search process, thus solving the collapse issue fundamentally. This decoder processes hierarchical features, integrating both fine and coarse-grained information, leading to improved reconstruction quality. Experimental results on ImageNet~\cite{deng2009imagenet} and CIFAR \cite{krizhevsky2009learning} dataset show that MAE-NAS outperforms existing methods while maintaining comparable complexity and adhering to the same search space constraints. Further comprehensive experimental analyses and ablation studies provide deeper insights into the robustness and characteristics of MAE-NAS as a NAS learner.

In summary, our contributions are as follows:

\begin{itemize}
\item We propose an innovative neural architecture search (NAS) method based on masked autoencoders, enabling efficient \emph{label-free searching} through self-supervised representation learning and directly addressing the challenge of resource-intensive labeled data requirements while maintaining architectural adaptability.
 \item Our approach is designed to be plug-and-play, integrating effortlessly with existing supervised NAS methods. And it is compatible with various DARTS variants, eliminating the need for handcrafted indicators and proving its integration capability without additional overhead.
\item MAE-NAS achieves superior performance with less search cost than its counterparts. Furthermore, it presents a novel perspective for addressing the performance collapse issue in DARTS under unsupervised scenarios.
\end{itemize}

\section{Method}

\subsection{DARTS Enhanced with Masked Autoencoders}
\label{sec:method}

Consider $\mathcal{L}_{train}$ and $\mathcal{L}_{val}$ as the training and validation losses, respectively. In the DARTS framework~\cite{liu2018darts}, the objective is to discover $\alpha^{*}$ that minimizes the validation loss $\mathcal{L}_{val}(w^{*}, \alpha^*)$, where the weights $w^*$ are acquired by minimizing the training loss $w^* = \arg\min_w \mathcal{L}_{train}(w, \alpha^*)$. This approach results in a two-level optimization problem:

\begin{equation}
    \begin{aligned}
    \mathop{\min}_{\alpha} \quad &\mathcal{L}_{val}(w^{*}(\alpha), \alpha). \\
    s.t. \quad & w^{*}(\alpha) = \arg\min_{w} \mathcal{L}_{train}(w, \alpha)
    \end{aligned}
    \label{darts_equation}
\end{equation}  

Our proposal, MAE-NAS, stems from a critical observation: supervised neural architecture search may overfit the training data. Despite optimizing both $\alpha$ and $w$ in Equation \ref{darts_equation}, these models tend to achieve near-zero errors on training sets, yet the real aim is to find architectures that generalize well on the validation set. This illustrates an inherent contradiction in supervised learning. To address this, we leverage SimMIM~\cite{xie2022simmim} as a proxy task for NAS, aiming to discover models with superior generalization in an unsupervised context. Extending DARTS, the new optimization goal is:

\begin{equation}
    \begin{aligned}
    \mathop{\min}_{\alpha} \quad &\mathcal{L}_{val}^{M}(w^{*}(\alpha), \alpha, M), \\
    s.t. \quad & w^{*}(\alpha) = \arg\min_{w} \mathcal{L}_{train}^{M}(w, \alpha^*, M)
    \end{aligned}
    \label{darts_mea_equation}
\end{equation}  
here $M$ is the set of masked pixels, while $\mathcal{L}_{val}^{M}$ and $\mathcal{L}_{train}^{M}$ are image reconstruction losses akin to SimMIM. The SSL-style objective functions as the exclusive loss function, incurring no extra training costs. Illustrated in Figure \ref{fig:mae-nas}, the method includes an encoder that converts the input into latent representation, alongside a decoder to reconstruct the original input. Utilizing the supernet of DARTS as the encoder backbone, the masked autoencoder evolves into a robust NAS learner, discovering promising architectures from the DARTS search space with minimal image reconstruction loss. Implementing MAE on convolution networks poses challenges; thus, we generally follow SparK~\cite{tian2023designing} for this purpose.

\begin{table*}[t]
\centering
\renewcommand{\arraystretch}{1.05}
\setlength{\tabcolsep}{3.0pt}
\small
\caption{Comparison of NAS methods on CIFAR-10 and ImageNet. We report the average of 5 runs.}
\label{tab:narrow}
\begin{tabular}{l|rrrrrrr}
\toprule
Datasets & Model & Params (M) & FLOPs (M) & Top-1 (\%) & Top-5 (\%) & Supervised & Cost (days) \\
\midrule
\multirow{9}{*}{CIFAR-10}
& NASNet-A~\cite{zoph2017learning} & 3.3 & 608 & 97.35 & -- & Yes & 2000\\
& ENAS~\cite{pham2018efficient} & 4.6 & 626 & 97.11 & -- & Yes & 0.5\\
& DARTS~\cite{liu2018darts} & 3.3 & 528 & 97.0 & -- & Yes & 0.4\\
& GDAS~\cite{dong2019searching} & 3.4 & 519 & 97.07 & -- & Yes & 0.2\\
& P-DARTS~\cite{chen2019progressive} & 3.4 & 532 & 97.5 & -- & Yes & 0.3\\
& PC-DARTS~\cite{xu2020pcdarts} & 3.6 & 558 & 97.43 & -- & Yes & 0.1\\
& ROME-v1~\cite{wang2023rome} & 4.5 & 683 & 97.5 & -- & Yes & 0.3\\
& DARTS-~\cite{chu2020darts} & 3.5 & 568 & 97.5 & -- & Yes & 0.4\\
& \textbf{Ours} & 3.6 & 561 & \textbf{97.8} & -- & No & 0.2\\[1mm]
\midrule
\multirow{12}{*}{ImageNet}
& NASNet-A~\cite{zoph2017learning} & 5.3 & 564 & 74.0 & 91.6 & Yes & 2000\\
& DARTS~\cite{liu2018darts} & 4.7 & 574 & 73.3 & 91.3 & Yes & 0.4\\
& SNAS~\cite{xie2018snas} & 4.3 & 522 & 72.7 & 90.8 & Yes & 1.5\\
& FairDARTS-B~\cite{chu2019fair} & 4.8 & 541 & 75.1 & 92.5 & Yes & 0.4\\
& AmoebaNet-A~\cite{real2019regularized} & 5.1 & 555 & 74.5 & 92.0 & Yes & 3150\\
& MnasNet~\cite{tan2018mnasnet} & 3.9 & 388 & 74.79 & 92.1 & Yes & 3791\\
& FBNet-C~\cite{wu2018fbnet} & 5.5 & 375 & 74.9 & 92.3 & Yes & 9\\
& FairNAS-A~\cite{chu2019fairnas} & 4.6 & 388 & 75.3 & 92.4 & Yes & 12\\
& PC-DARTS~\cite{xu2020pcdarts} & 5.3 & 597 & 75.8 & 92.7 & Yes & 3.8\\
& SWAP-NAS~\cite{peng_2024} & 5.8 & -- & 76.0 & 92.4 & No & 0.006\\
& UnNAS~\cite{liu2020labels} & 5.1 & 552 & 75.8 & - & No & 4.8\\
& RLNAS~\cite{zhang2021neural} & 5.2 & 561 & 75.9 & - & No & 8.33 \\
& \textbf{Ours} & 4.9 & 547 & \textbf{76.5} & \textbf{93.3} & No & 2.3\\
\bottomrule
\end{tabular}
\end{table*}

\subsection{Hierarchical Decoder}
\label{sec:HD}
DARTS suffers from performance collapse when skip connections dominate in supervised settings, attributed to unfair operator competition during supernet training~\cite{chu2020darts,xu2020pcdarts}. While regularization methods like R-DARTS~\cite{zela2020understanding} mitigate this through $L_2$  or \emph{ScheduledDropPath}, we investigate if unsupervised paradigms face similar risks. Our experiments based on MAE-NAS reveal collapse correlates with mask ratio: ratios below 0.5 induce collapse, while below 0.5 prevent it. We attribute this to MAE's inherent regularization: larger mask ratios intensify reconstruction difficulty, forcing encoders to discover robust architectures rather than settling for suboptimal shortcuts. This aligns with supervised conclusions: the mask ratio effectively controls regularization strength, mirroring explicit methods in supervised NAS.

Adjusting mask ratios risks excluding promising architectures via threshold sensitivity. Addressing DARTS' collapse from unfair competition and instability~\cite{chu2019fair}, our hierarchical decoder processes multi-resolution features $F_1$, $F_2$, and $F_3$ from the encoder, reconstructing them through parallel paths:

\begin{equation}
    \begin{split}
    & I_{rec} = Linear(Conv(F_1) +  \\ & Upsample(Conv(F_2), 2) + Upsample(Conv(F_3), 4)).
    \end{split}
\end{equation}
Loss focuses only on masked patches in $I_{rec}$ . Unlike SimMIM's transformer-based reconstruction, our lightweight decoder (equivalent to one conv layer) preserves DARTS' computational efficiency while enforcing multi-scale learning.

The hierarchy delivers dual benefits: accelerated gradient flow with shallower paths, and inherent stabilization from multi-branch interactions. Crucially, it aligns with convnets' multi-scale paradigm, unlike transformer-centric masked modeling. For example, pyramid structures~\cite{lin2017feature} prove hierarchical decoding essential for vision tasks where single-scale features fail to handle object size variations. Thus, the hierarchical decoder design makes our approach particularly suited to convnet-based NAS spaces.

\subsection{Relationship to Prior Works}

Label-free NAS is well-established, with studies like UnNAS~\cite{liu2020labels} and RLNAS~\cite{zhang2021neural} demonstrating its comparable effectiveness to supervised NAS. However, our work is the first to explore the Masked Autoencoder (MAE) paradigm in unsupervised NAS, which presents unique challenges. A naive MAE implementation leads to performance collapse similar to DARTS issues, addressed by our novel hierarchical decoder integrated with the masked autoencoder objective. Extensive evaluations across diverse benchmarks confirm the robustness of our approach. The MAE proxy proves more efficient and versatile than existing unsupervised metrics. Current alternatives face limitations: RLNAS's angle metric fails with non-parametric operators like mixed activation/pooling layers, while UnNAS proxies exhibit dataset-dependent inconsistency~\cite{liu2020labels}. In contrast, our method overcomes these constraints while maintaining universal applicability.

\section{Experiments}
\subsection{Comparisons with State-of-the-art Methods}

We conduct extensive experiments on DARTS search space, following the experimental protocols of DARTS- \cite{chu2020darts}. For mask image modeling, the mask ratio is set to 0.5, with patch sizes of 8 and 4 for all datasets. 

\noindent\textbf{Supervised NAS comparison}. Despite operating without labels, As in Table \ref{tab:narrow}, MAE-NAS achieves 97.8\% and 76.5\% top-1 accuracy  while requiring only 0.2 and 2.3 GPU days on CIFAR-10 and ImageNet datasets respectively, which outperforms all supervised counterparts with a clear margin. 

\noindent\textbf{Unsupervised NAS comparison}. Following the standard unsupervised NAS paradigm \cite{liu2020labels,zhang2021neural}, Table \ref{tab:narrow} demonstrates MAE-NAS's advantages over UnNAS and RLNAS in both performance and efficiency. Our method achieves the best accuracy with reduced FLOPs and the lower search costs. Since zero-shot NAS requires no training during the search phase, its search cost is significantly lower, which is fundamentally a distinct paradigm. While progress in the mature DARTS search space remains challenging due to extensive prior optimization, our results highlight masked autoencoders' potential as effective proxies for NAS tasks.

\subsection{Sensitivity Analysis of Mask Ratio and Patch Size}

Mask ratio (proportion of obscured pixels) and patch size (spatial scale of masked regions) critically influence MAE's ability to learn robust representations through image reconstruction. We systematically evaluate these parameters in MAE-NAS by varying mask ratios (0.1–0.7, step=0.2) and patch sizes (2–16, step=2). As shown in Table \ref{tab:cifar_10_mask_ratio_and_path}, both parameters exhibit minimal impact on architecture search outcomes, demonstrating our method's robustness. While MAE lacks built-in hyperparameter adaptation, empirical results confirm that default settings effectively identify high-performing architectures without requiring parameter tuning.

\begin{table}[h]
    \setlength\tabcolsep{1pt}
    \centering
    \begin{tabular}{cc|cc}
        \toprule
        Mask Ratio & Top-1 Error (\%) & Patch Size & Top-1 Error (\%) \\
        \midrule
            0.1 & 2.79$\pm$0.18 & 2 & 2.75$\pm$0.24\\
            0.3 & 2.77$\pm$0.14 & 4 & 2.71$\pm$0.16\\
            0.5 & 2.75$\pm$0.09 & 8 & 2.71$\pm$0.12\\
            0.7 & 2.78$\pm$0.16 & 16 & 2.74$\pm$0.21\\
        \bottomrule
    \end{tabular}
    \caption{Search performance on CIFAR-10 in S0 with respect to mask ratio and patch size. Each setting is run three times.}
    \label{tab:cifar_10_mask_ratio_and_path}
\end{table}

\subsection{Ablation of Hierarchical Decoder}
To assess the impact of the hierarchical decoder (HD), we replaced the HD module with a conventional decoder as used in SimMIM \cite{xie2022simmim}. Table \ref{tab:collapse_statistic} displays the number of skip connections and the top-1 accuracy of architectures identified in both configurations. It is apparent that MAE-NAS without HD is significantly affected by performance collapse when the mask ratio is relatively low (i.e., below 0.5), whereas our method maintains stable performance across diverse mask ratios.
\begin{table}[h]
    \setlength\tabcolsep{1pt}
    \centering
    \begin{tabular}{c|cccc}
        \toprule
        & Mask Ratio & Top-1 Error (\%) & No. of skips \\
        \midrule 
        Ours  &  0.2 & 2.67$\pm$0.14 & 1\\  
            & 0.4 & 2.65$\pm$0.16 & 1 \\    
            & 0.6 & 2.68$\pm$0.15 & 1 \\ 
            & 0.8 & 2.66$\pm$0.21 & 1 \\
        \midrule
        w/o HD & 0.2 & 5.87$\pm$0.59 & 8 \\
            & 0.4 & 4.23$\pm$0.37 & 6 \\ 
            & 0.6 & 2.87 $\pm$0.29 & 2 \\
            & 0.8 & 2.67$\pm$0.22 & 1 \\ 
        \bottomrule
    \end{tabular}
    \caption{Performance evaluation on CIFAR-10 with varying mask ratios. HD denotes Hierarchical Decoder. Each experiment is repeated three times.}
    \label{tab:collapse_statistic}
\end{table}

\subsection{Visualization of Image Reconstruction}
\begin{figure}[h]
  \centering
  \includegraphics[width=0.47\textwidth]{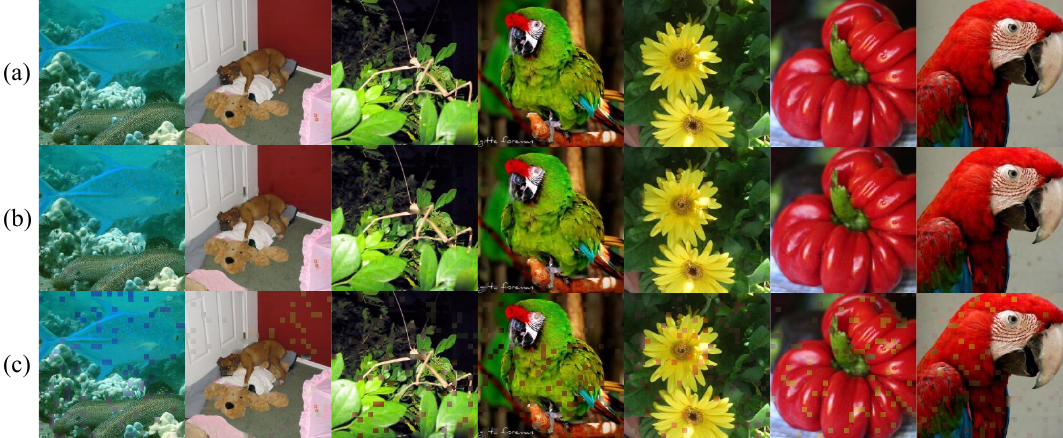}
  \caption{Comparison of the original images (a) and the reconstructed images on ImageNet. The second and third rows represent the reconstructed images of the MAE-NAS supernet under the settings of w/ HD (b) and w/o HD (c) respectively. }
  \label{fig:reconstruction_img}
\end{figure}

In Section \ref{sec:HD}, we explain how HD prevents performance collapse by accelerating gradient propagation and improving training stability. This analysis extends to image reconstruction: Figure \ref{fig:reconstruction_img} shows MAE-NAS achieves better reconstruction quality than non-HD counterparts. Effective architectures emerge naturally through the encoder's need to restore masked images accurately, and the superior reconstruction quality promotes learning powerful architectures. This mechanism inherently prevents convergence on inferior architectures, helping MAE-NAS avoid performance collapse.

\section{Related Work}
Manually designed neural networks excel in computer vision, but their architectures are likely sub-optimal, fueling rising NAS research in academia and industry.

\noindent\textbf{Supervised NAS.} Supervised NAS has become a leading paradigm. Weight-sharing NAS is a key branch of NAS, which reduces computational costs by two approaches~\cite{liu2018darts,chu2020darts}: (1) One-shot supernet training with candidate evaluation; (2) Gradient-based joint optimization of architecture. Our method extends DARTS' differentiable framework.

\noindent\textbf{Unsupervised NAS.} Recently, there has been increasing interest in applying unsupervised learning to NAS, driven by its potential to reduce the reliance on labeled data~\cite{zhang2021neural,liu2020labels,peng_2024}. However, no studies have yet explored the use of masked autoencoders for fully unsupervised NAS.

\section{Conclusion}
Acquiring labeled data is time-consuming, highlighting the appeal of unsupervised Neural Architecture Search (NAS). We propose MAE-NAS, which uses Masked Autoencoders to eliminate labeled data requirements by replacing supervised objectives with reconstruction loss, improving search efficiency and architectural quality. Particularly, the hierarchical decoder is proposed for addressing the issue of performance collapse in DARTS. Our experiments show MAE-NAS outperforms existing methods under the same even less complexity constraints. While experiments focus on image classification, applying this approach to image generation requires further exploration.

\vfill\pagebreak
\bibliographystyle{unsrt}
\small\bibliography{example_paper}

\end{document}